\documentclass[sigconf]{aamas} 

\usepackage{balance} 
\usepackage{fancyvrb} 
\usepackage{tabularx}
\newcommand{\edit}[1]{{#1}}
\newcommand{\revise}[1]{{#1}}
\newcommand{\pascal}[1]{{#1}}

\setcopyright{ifaamas}
\acmConference[AAMAS '23]{Proc.\@ of the 22nd International Conference
on Autonomous Agents and Multiagent Systems (AAMAS 2023)}{May 29 -- June 2, 2023}
{London, United Kingdom}{A.~Ricci, W.~Yeoh, N.~Agmon, B.~An (eds.)}
\copyrightyear{2023}
\acmYear{2023}
\acmDOI{}
\acmPrice{}
\acmISBN{}

\acmSubmissionID{316}

\title{FedFormer: Contextual Federation with Attention in Reinforcement Learning}

\author{Liam Hebert}
\affiliation{
  \institution{University of Waterloo}
  \city{Waterloo}
  \country{Canada}}
\email{liam.hebert@uwaterloo.ca}

\author{Lukasz Golab}
\affiliation{
  \institution{University of Waterloo}
  \city{Waterloo}
  \country{Canada}}
\email{lgolab@uwaterloo.ca}

\author{Pascal Poupart}
\affiliation{
  \institution{University of Waterloo, Waterloo, Canada}
  \city{Vector Institute, Toronto}
  \country{Canada}}
\email{ppoupart@uwaterloo.ca}

\author{Robin Cohen}
\affiliation{
  \institution{University of Waterloo}
  \city{Waterloo}
  \country{Canada}}
\email{rcohen@uwaterloo.ca}

\begin{abstract}
A core issue in multi-agent federated reinforcement learning is defining how to aggregate insights from multiple agents. This is commonly done by taking the average of each participating agent's model weights into one common model (FedAvg). We instead propose FedFormer, a novel federation strategy that utilizes Transformer Attention to contextually aggregate embeddings from models originating from different learner agents. In so doing, we attentively weigh the contributions of other agents with respect to the current agent's environment and learned relationships, thus providing a more effective and efficient federation. We evaluate our methods on the Meta-World environment and find that our approach yields significant improvements over FedAvg and non-federated Soft Actor-Critic single-agent methods. Our results compared to Soft Actor-Critic show that FedFormer achieves higher episodic return while still abiding by the privacy constraints of federated learning. Finally, we also demonstrate improvements in effectiveness with increased agent pools across all methods in certain tasks. This is contrasted by FedAvg, which fails to make noticeable improvements when scaled.  
\end{abstract}


\newcommand{\BibTeX}{\rm B\kern-.05em{\sc i\kern-.025em b}\kern-.08em\TeX}

\begin{document}


\pagestyle{fancy}
\fancyhead{}


\maketitle 


\section{Introduction}

Reinforcement learning has become a ubiquitous tool 
in applications such as autonomous vehicles \citep{Sallab_2017, shalev2016Safe}, where vehicles learn to navigate the streets of many complex environments, and Internet-of-Things devices \citep{mohammadi2017semisupervised, bu2019smart, xiong2020resource}. However, to obtain robust performance,  
a large collection of training episodes is often needed. In the case of safety-critical systems such as autonomous vehicles, training episodes must also be diverse enough to account for the many possible edge cases that the system could face. As a result, the scale and diversity of data collection needed to achieve adequate performance quickly becomes infeasible for a single party. To account for these challenges, a common technique for data collection relies on crowdsourcing, where training episodes are collected from many agents that explore their local environment and then transmitted to a central controller to train a generalized model \citep{qi2021federated}. 

However, this strategy comes with a crucial challenge: privacy. In our example of autonomous vehicles, training data consist of sensitive images of local surroundings, which need to be broadcast over the internet to a centralized source. This presents an opportunity for bad actors to intercept these data for malicious purposes. Motivated by this challenge, privacy-preserving federated machine learning and federated reinforcement learning have been proposed. Here, rather than transmitting training observations, local model weights are transmitted instead. These model weights are then aggregated by a centralized controller to form a new common model, all without seeing any sensitive data.

The \edit{current state-of-the-art general} method in federated \edit{reinforcement} learning to combine agents is FedAvg \citep{mcmahan2017communication}. This strategy takes the weighted average of transmitted model weights across all agents according to the number of observations that an agent has seen. However, there are several drawbacks that come with this approach. In reinforcement learning, policy and critic networks are already prone to instability, which could be further exacerbated by the averaging of model parameters. In addition, the averaging of model weights across multiple agents often prohibits individual exploration in non-identical environments, a challenge when federating models explore diverse environments \citep{qi2021federated}. An example of this can be seen in autonomous vehicles, where knowledge gained from agents exploring rural roads is less useful to agents exploring urban cities. Despite this, contributions from both sets of agents would still be equally weighted under FedAvg, potentially creating a final agent that performs worse in both environments.

To address these drawbacks, inspired by recent advancements in natural language processing, we propose Federated Transformers (FedFormer), a novel federation policy based on transformer encoder models \citep{vaswani2017attention, devlin2018bert}. Instead of taking the average of model weights, we utilize transformer encoders to learn contextual relationships between agents. We then leverage these learned relationships to contextually federate agents together during inference. This method allows local agents to maintain models specific to their environment, allowing for local exploration, while also contextually including insights from other agents according to their relevance to the current environment. Importantly, these relationships can be learned without divulging confidential metadata, such as geolocation data, health information or other sensitive attributes. In addition, we compute these relationships during each time step, allowing for the dynamic federation as an agent's environment changes throughout the episode. This allows our approach to providing a more efficient aggregation of model insights during inference.  

We evaluate FedFormer using the Meta-World environment \citep{yu2020meta}, a benchmark of robotic manipulation tasks.
Despite training on heterogeneous environments, our approach brings improved episodic return over FedAvg and over non-federated Soft Actor-Critic (SAC) \citep{haarnoja2018soft} single-agent methods. This is despite SAC being trained on the entire set of environments as the federated strategies, representing performance without the privacy constraints of the federation. \edit{We also demonstrate the onboarding performance of FedFormer, finding that convergence speed increases 5-fold by leveraging pre-trained agents in the federation network despite being applied to never-before-seen environments.} Notably, these results are obtained without intervention from the other agents, further increasing privacy and effectiveness. When scaled to 10 and 15 agents, FedFormer outperforms the episodic return of FedAvg by a factor of 2.4 and 3.8 at the last epoch,  respectively. This illustrates that FedFormer performs well when scaled to increased agent pools (with matching or increased performance), whereas FedAvg fails to make noticeable improvements or presents degraded performance when scaled. In all, our results illustrate that FedFormer performs better than \edit{state-of-the-art} FedAvg, scales without degradation of performance, and is more effective than single-agent Soft Actor-Critic while still abiding by the privacy constraints of federated learning. Our source code is available at \href{https://github.com/liamhebert/FedFormer}{https://github.com/liamhebert/FedFormer}.

\section{Background}
\label{sec:background}
\subsection{Federated Learning and Federated Reinforcement Learning}
There are four key properties of federated learning: distribution, data protection, generality and status equality \citep{qi2021federated}. Distribution states that federated model training is done in parallel by all participating models. Data protection ensures that the training data held by each participating party is not transmitted to other parties. Generality states that federated models must generalize to diverse environments. Finally, status equality ensures that training favours all participating parties equally, and often with identical infrastructures. 

Satisfying these constraints, \citet{mcmahan2017communication} proposed the FedAvg federation strategy. Given a set of $N$ parties with loss functions $\{\mathcal{F}_i\}^N_{i=1}$ and datasets $\{D_i\}^N_{i=1} \in D$ interested in formulating a joint cooperative model, FedAvg updates a centralized model as 
\begin{align}
    \forall i, w_i(t + 1) &= w_i(t) - \gamma \nabla F_i(w_i(t)) \\
    w_g(t + 1) &= \sum_{i = 1}^N \frac{|D_i|}{|D|} w_i(t + 1)
\end{align}
where $w_i(t)$ is the model update computed by agent $i$ at time step $t$, $\gamma$ is a fixed learning rate and $w_g$ are the parameters of the centralized model. Under this policy, one can tune how often each model communicates its parameters to the centralized source. This strategy has been transferred to reinforcement learning, with many techniques applying FedAvg to federate policy and critic networks \citep{qi2021federated}. The most common extension of FedAvg is to introduce domain-specific adjustments to the weighing of model parameters. This has been done in the context of the game Pong \citep{nadiger2019federated}; for ioT devices, to optimize edge computation  \citep{jianji2019federated, wang2020federated, tehrani2021Federated}, and for the valued, recent application of real-time electric vehicle charging
\citep{zhang2022federated}. 

Another direction in federated reinforcement learning is to utilize metadata about agents to weigh their contributions. \revise{\citet{lim2020federated} and \citet{chu2022multiagent}
weigh agent contributions according to their average reward.} 
Similarly, \citet{wang2020attention} propose weighing FedAvg according to metadata such as batch size, average loss, dataset size and hit rate. In each of these prior settings, the training environments are similar, allowing for a non-disruptive application of parameter averaging. \revise{We refer to this class of federated methods as Weighted Federated Averaging.}


However, it has also been shown that the performance of FedAvg degrades significantly when environments are perturbed and are heterogeneous between agents \citep{qi2021federated}. This is important since in many prior works, it is assumed that the environments of agents are consistent and non-heterogeneous \cite{jianji2019federated, wang2020federated, tehrani2021Federated}. For other examples of federated reinforcement learning, we refer the reader to a survey by \citet{qi2021federated}. Since most of the prior work relies on FedAvg with minor adjustments towards domain-specific goals, different model structures or additional processing steps prior to applying FedAvg, we focus our attention on directly comparing to FedAvg and Weighted FedAvg. This aligns with our focus on proposing a method that can directly replace FedAvg with a focus on applications for federated reinforcement learning. 

In this work, we mitigate the downfalls of FedAvg \revise{and propose a general framework for federated reinforcement learning that utilizes an attention-based aggregation mechanism to federate agent predictions.} Instead of using a hand-made federation strategy, we utilize a transformer model to learn contextual relationships between agents for federation. This approach allows federated reinforcement learning to be applied to tasks where environments are significantly different while still benefiting from the mutual federation. 

\subsection{Reinforcement Learning}

\pascal{Reinforcement Learning (RL) is based on a formalization of Markov Decision Processes with a set $S$ of states $s$, a set $A$ of actions $a$, a transition function $P(s'|s,a)$ that indicates the probability of reaching state $s'$ when executing action $a$ in state $s$, a reward function $R(s,a)=r\in\Re$ that returns the expected immediate reward of executing $a$ in $s$, a discount factor $\gamma \in [0,1)$, and a planning horizon $h$ (assumed to be infinite throughout this paper).  In RL, an agent optimizes a policy $\pi(a|s)$ that indicates the probability of executing $a$ in $s$. The value $J^\pi(s)$ of executing a policy $\pi$ when starting in state $s$ is the sum of expected discounted rewards (i.e., $J^\pi(s_0)=E_\pi[\sum_{t=0}^h \gamma^t R(s_t,a_t)]$). Similarly, the value of executing $a$ in $s$ followed by the policy $\pi$ is $Q^\pi(s_0,a_0)=R(s_0,a_0)+\gamma E_\pi[\sum_{t=1}^h \gamma^t R(s_t,a_t)]$. An optimal policy $\pi^*$ achieves the highest expected cumulative reward (i.e., $\pi^* = argmax_\pi J^\pi(s) \forall \pi,s$). However, since the transition distribution and the reward function are unknown, the agent must perform this optimization based on trajectories of states, actions and rewards (i.e., $s_1, a_1, r_1, s_2, a_2, r_2, ...$) that it experiences as it interacts with the environment. }  

\subsubsection{Soft Actor-Critic (SAC)}
\label{sec:softactor}
Within reinforcement learning, traditional policy gradient techniques update the parameters $\theta$ of a policy $\pi_\theta$ by taking steps in the direction of the gradient of the value function \citep{sutton1999policy, williams1992simple}:  
\begin{align}
   \nabla _\theta J^{\pi_\theta}(s_t) = \nabla_\theta log(\pi_\theta(a_t | s_t)) \sum^\infty_{t' = t} \gamma^{t' - t}R(s_{t'}, a_{t'}) 
\end{align}

However, a key weakness of this approach is the high variance brought on from the sum of discounted rewards for a given episode $\sum_{t' = t}^\infty$, as rewards can vary significantly from episode to episode \citep{konda1999actor}. To account for this, Actor-Critic methods were proposed, which utilize a function to approximate the sum of expected rewards and train that function separately. This function, called the Q-function or critic function, is trained through off-policy temporal-difference learning by minimizing:
\begin{align}
    \mathcal{L}_Q(\psi) &= \mathbb{E}_{(s, a, r, s')\approx D}(Q_\psi(s, a) - y)^2 \\
    y &= r(s, a) + \gamma \mathbb{E}_{a' \approx \pi(s')} Q_{\Hat{\psi}(s', a')}
\end{align}
where $Q_{\Bar{\psi}}$ is a target Q-function (parameterized by $\psi$), an average of past Q-functions, and $D$ is a replay buffer of past episodes. 
Recent approaches towards Actor-Critic methods include an entropy term during the policy gradient step to encourage exploration \citep{haarnoja2018soft}. This new variant, titled Soft Actor-Critic (SAC), proposes a soft policy update given as 
\begin{align}
    & \nabla_\theta J(\pi_\theta) \\ & = \mathbb{E}_{s\approx D, a \approx \pi} \nabla_\theta log(\pi_\theta(a | s) (- \alpha log(\pi_\theta(a | s)) + Q_\psi(s, a) - b(s))) \nonumber
\end{align}
where $b(s)$ is a learned state-dependent baseline function. 

\section{Methods}
\label{sec:arch}
\begin{figure}
    \centering
    \includegraphics[width=\linewidth]{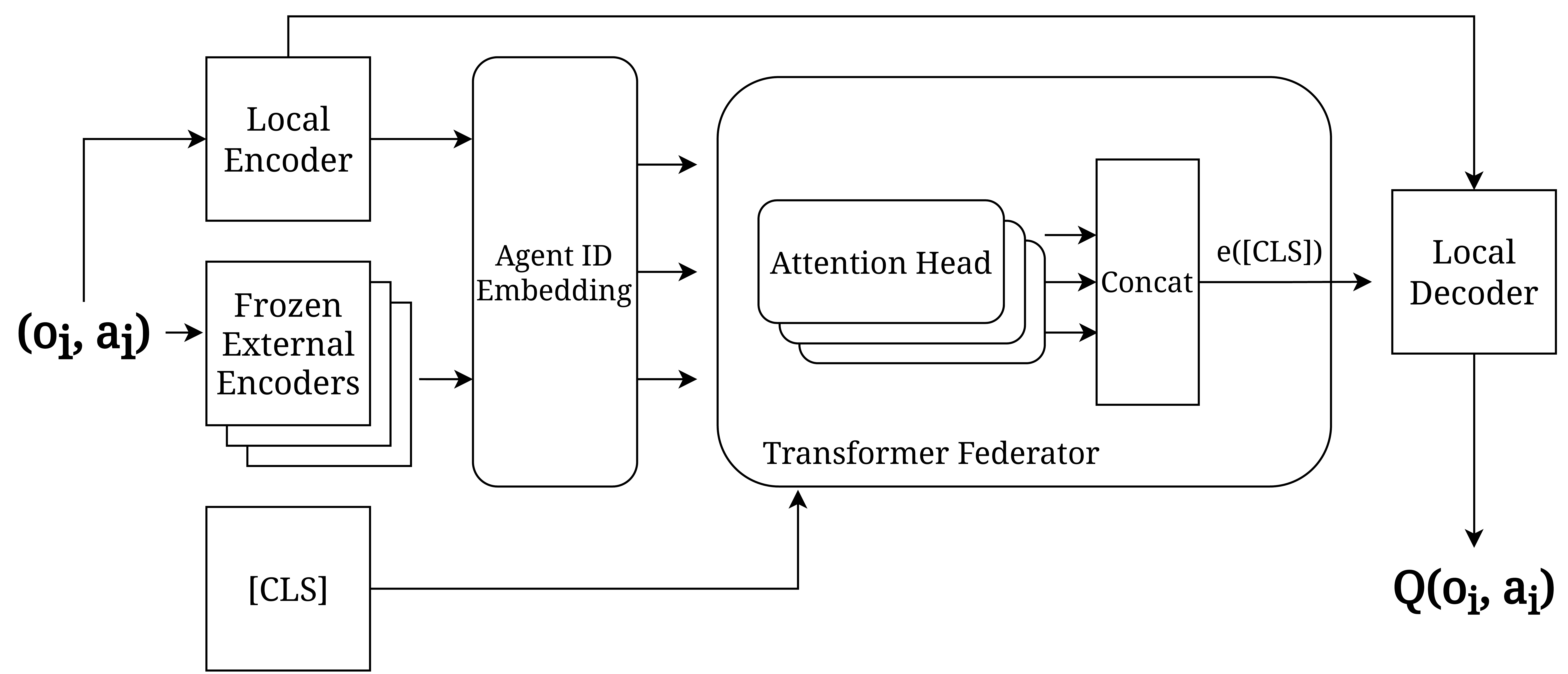}
    \caption{Federated Transformer Q-Function}
    \label{fig:arch}
\end{figure}
Our FedFormer model adapts the SAC algorithm by introducing a contextual federated Q-function (Section \ref{sec:softactor}). Our only core modification from standard Soft Actor-Critic is the usage of double Q-Networks to help model stability \citep{fujimoto2018addressing}. 

The overall architecture for our federated Q-function can be seen in Figure \ref{fig:arch}. Given a network of $N$ agents, we initialize the Q-network of each agent with $N - 1$ frozen external encoder networks, each representing the other agents, and one local encoder network. We also initialize a local transformer encoder and a final output decoder network. Each encoder network consists of the same infrastructure, with the exception of each external network having gradients disabled. 

The first step during inference is to generate encodings from each encoder network for the same action-observation pair. Following inspiration from BERT, we first encode agent identity through element-wise added learned embeddings according to the encoder network that generated that representation \citep{devlin2018bert} (IdEmbedding). This allows the aggregator network to identify the source of each encoding to potentially aid in learning which agents are known to be more relevant to the current agent. We also include a special \Verb$[CLS]$ embedding to the input set to encode the global representation of the transformer network. This stage of encodings $E$ is thus given as 
\begin{align}
   E = \{FF_i(o, a) + IdEmbedding(i), \forall i \in N, CLSEmbedding\}
\end{align}
where $FF_i$ is the encoder network for agent $i$. We then feed our set of encodings $E$ into a transformer encoder network. Given this set of embeddings, each layer of our transformer encoder network computes:
\begin{align} 
    Q = EW_Q, ~K = EW_K, ~V = EW_V, \\
    Attn(E) = softmax(\frac{QE^\top}{\sqrt{d}})V
\end{align}
where $Attn(E)$ is an attention-weighted representation of E and $W_Q, W_K, W_V$ are all learnable weight matrices representing the Query, Key and Value embeddings of the input. In our work, we utilize a two-layer transformer encoder architecture, but this can be tuned in future work.  

After aggregating each representation through our transformer encoder, we concatenate the transformed encoding of the \Verb$[CLS]$ token ($e_{[CLS]}$) with the local encoder representation $FF_l((o, a))$ into a final decoder network, predicting the Q-value for that action-observation pair for agent $i$. This is given as:
\begin{align}
    Q_i(o, a) = FF_d(e(\texttt{[CLS]}), FF_{local}(o, a))
\end{align}

Since each agent has its own transformer encoder network that is trained according to the agent's own local loss function, the \Verb$[CLS]$ encoding is therefore trained to be a global representation relevant to the task of that agent. During back-propagation, we fine-tune end-to-end apart from the external encoder networks.

At the end of every epoch, the frozen external networks are updated with the local encoder networks from the other agents. Since each agent needs only to transmit the weights of its local encoder network to the other agents, network bandwidth is increased compared to a FedAvg-based strategy, but not proportionally due to the increase in model complexity. In practice, network bandwidth is kept negligible due to the small size of local encoder networks. For our experiments, each agent was required to upload 1.8MB and download $1.8\text{MB} \times N-1$, which is negligible under modern networks. In addition, we only replace the encoders in the Q-function, not the target Q-function, as the target Q-function is eventually replaced by the Q-function regardless as part of Soft Actor-Critic \citep{haarnoja2018soft}.  

An interesting aspect of our architecture is the ease of agent onboarding, the process by which agents are introduced into the federation network. Since transformer networks can scale to a variety of input sizes, onboarding new agents consist of simply adding a new encoder network to the set of external encoder networks. This is significantly different from FedAvg, which could introduce untrained noise into other networks through parameter averaging. 

\section{Results}
\label{sec:exp}

\subsection{Experimental Design}
\label{sec:design}
To evaluate FedFormer, we utilize the Meta-World benchmark \citep{yu2020meta}, which includes a set of robotic manipulation tasks, each able to be parameterized with random variables. As a result, we can initialize a wide variety of diverse heterogeneous environments while maintaining the same task, a challenge for traditional FedAvg methods. For this work, we evaluate our methods on a set of seven tasks that concern operating a robotic arm to push buttons, open and close windows, doors, and drawers as well as reach specific destinations, among other complex tasks. \footnote{While the Meta-World benchmark can assess how well a meta-learning agent improves over a collection of tasks, we instead evaluate how well our solution performs on each of the several different tasks individually.}

\revise{We compare our solution against Federated Averaging (FedAvg) \cite{mcmahan2017communication}, Weighted FedAvg (FedWeightedAvg) as implemented in \citet{chu2022multiagent} and non-federated Soft Actor-Critic (SAC) \cite{haarnoja2018soft}}. As discussed in Section~\ref{sec:background}, recent advances in federated reinforcement learning are grounded in federated averaging with many domain-specific adjustments. Therefore, we focus our evaluation on comparing underlying federation strategies as our method can behave as a direct replacement for FedAvg. We also compare our methods against SAC to measure performance degradation by preserving privacy in a federated setting. Our experiments against SAC are trained on a centralized set of environments used by the federation methods. It is important to note that performance degradation is expected due to the restrictions of privacy preservation. As such, matching or exceeding non-federated methods is an extremely difficult benchmark and is often referred to as an upper limit for federated learning \citep{qi2021federated}. 

In each experiment, we follow the same hyper-parameters as proposed for SAC in the Meta-World paper \cite{yu2020meta}, with the exception that we train for 250 epochs with 600 training iterations during each epoch due to computing limitations. We utilize a batch size of 1200 steps and keep a replay buffer of size $10^6$. Each agent comprising the evaluated federation strategies learns from a set of five sampled environment configurations and is tested on a different set of five sampled environments. As such, each agent has access to a small constrained set of heterogeneous environments. Both federation strategies (FedAvg and our FedFormer) have five participating agents unless otherwise stated. For non-federated SAC, the single agent learns from the set of all sampled environments: 25 environments for training and 25 environments for testing. To ensure the robustness of our experiments, we utilize ten random seeds for each method and average the results with error bars reflecting standard error, represented as shaded areas in the figures. 

\begin{table}[t]
    \centering
    \caption{Hyperparameters Used}
    \begin{tabular}{l|l}
        \textbf{Parameter} & \textbf{Value}  \\
        \hline\hline
        Batch Size & 1200\\
        Number of Epochs & 250\\
        Path Length & 500\\ 
        Gradient Steps per Epoch & 600 \\
        Discount Factor & 0.99\\
        Optimizer & Adam \\
        Policy Hidden Sizes & (256, 256, 256) \\
        Policy Activation Function & ReLU \\
        Policy Learning Rate & $3 \times 10^{-4}$ \\
        Policy Minimum Standard Deviation & $e^{-20}$ \\
        Policy Maximum Standard Deviation & $e^{2}$ \\
        Soft Target Interpolation & $5 \times 10^{-3}$ \\
        Use Automatic Entropy Tuning & True \\
        \hline
        FedFormer \\
        \hline
        Transformer Layers & 2 \\
        Transformer Heads & 4 \\ 
        Transformer Hidden Sizes & 256 \\
        Encoder Hidden Size & (256, 256, 256) \\
        Decoder Hidden Size & (256, 256, 256) \\ 
        Batch Normalization & True \\
        Activation Function & ReLU \\
        Q-Function Learning Rate & $3 \times 10^{-4}$ \\
        \hline 
        SAC and FedAvg \\
        \hline
        Hidden Size & (256, 256, 256) \\
        Activation Function & ReLU \\
        Q-Function Learning Rate & $3 \times 10^{-4}$ \\
    \end{tabular}
    \label{tab:gen_hyper}
\end{table}


In each experiment, we use a three-layer feed-forward network with two output heads to predict the mean and variance of a tanh distribution over the action space as the policy network. For FedFormer, our Q-network consists of a two-layer transformer model with a hidden size of 400. Each encoder and decoder network consists of a three-layer feed-forward network with each layer having a hidden size of 400 except for the final decoder layer, which consists of a single output unit representing the Q-value. For FedAvg and SAC, the Q-network consists of a three-layer feed-forward network each with a hidden size of 400, apart from the last layer which has a single output unit representing the Q-Value. We include a detailed list of all hyperparameters in Table \ref{tab:gen_hyper}.

To implement our work, we modified an implementation of SAC in RLKit \citep{rlkit} with additional modules borrowed from Garage \citep{garage} and PyTorch \citep{paszke2019pytorch}. RLKit and Garage are licensed under the MIT license\footnote{https://spdx.org/licenses/MIT.html} and Pytorch is licensed under the BSD-3 "Revised" license \footnote{https://github.com/pytorch/pytorch/blob/master/LICENSE}. Experiments were run in parallel on a computing cluster consisting of RTX 3080, RTX 2080Ti, V100 and P100 GPUs, depending on availability. 

\subsection{Meta-World Evaluation}
\label{sec:performance}
\begin{figure*}
    \centering
    \includegraphics[width=\linewidth]{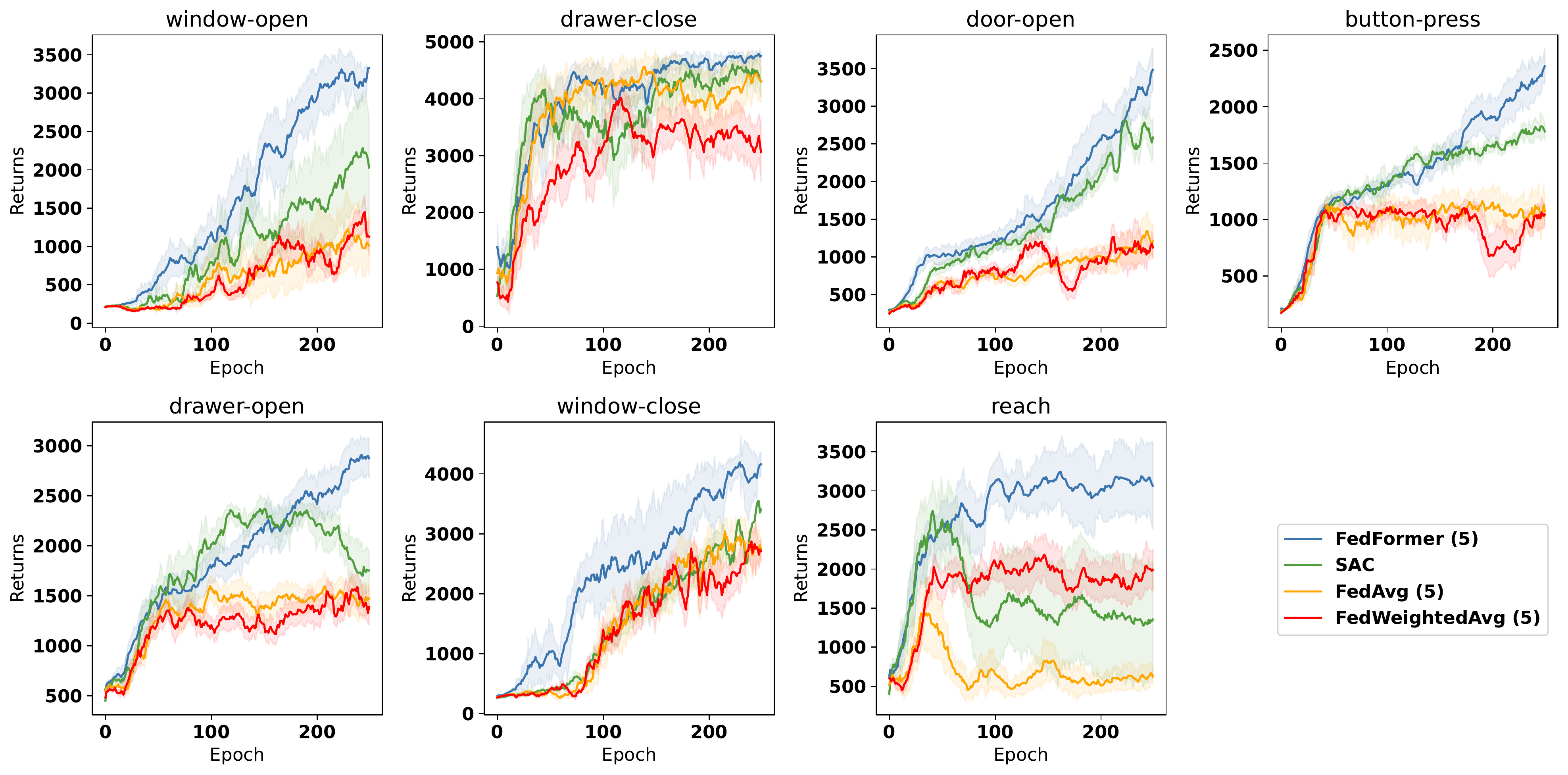}
    \caption{\revise{Performance on Meta-World tasks}}
    \label{fig:overall_performance}
\end{figure*}

In the first set of experiments, \revise{we compare the performance of FedFormer (blue) against FedAvg \citep{mcmahan2017communication} (orange), Weighted FedAvg \citep{chu2022multiagent} (red) and non-Federated SAC \citep{haarnoja2018soft} (green) on seven Meta-World tasks}. Figure \ref{fig:overall_performance} shows the results, where the X-axis represents training epochs and the Y-axis is the average episodic return. Improvements in the X-axis indicate that the agent is able to learn faster, and the Y-axis indicates enhanced performance, where the agent outperforms other agents. Further, separation in standard error (shaded area) between each method indicates improved performance based on 10 evaluated seeds. 

FedFormer outperforms FedAvg and Weighted FedAvg and matches or exceeds the performance of non-federated SAC in all of our evaluated tasks. As demonstrated by the standard error bars, FedFormer demonstrates consistently superior performance to FedAvg across six of the seven evaluated tasks. The only exception is the \texttt{drawer-close} task, in which all methods achieved similar returns.

The performance difference between FedFormer and FedAvg increases as training progresses. At the final epoch, FedFormer achieves 2.17x (\texttt{door-open}), 1.77x (\texttt{drawer-open}), 1.48x (\texttt{window-close}), 2.25x (\texttt{button-press}), 3.41x (\texttt{reach}) and 2.60x (\texttt{window-open}) performance gains over FedAvg. The only task where FedAvg performs comparably to FedFormer is on the \texttt{drawer-close} task. \revise{We also see that FedAvg behaves similarly to FedWeightedAvg with the exception of \texttt{window-close}, where FedWeightedAvg achieves much stronger performance.}

Notably, increases in effectiveness can be seen in the \texttt{button-press}, \texttt{drawer-open} and \texttt{door-open} tasks, where FedAvg appears to plateau early in performance around epoch 50. This is contrasted by FedFormer, which continues to grow in performance. We hypothesize that this can be attributed to stifled exploration in FedAvg, a key feature of reinforcement learning that differs from traditional machine learning. When an individual agent makes considerable gains in FedAvg, it is forced to average its parameters with those of the other agents, which could be performing much worse. This is overcome by FedFormer, which allows for attentive and selective aggregation between agents according to their learned relevance.  

Most surprising are the results compared to non-federated SAC. In the \texttt{door-open}, \texttt{reach}, \texttt{window-open} and \texttt{window-close} tasks, FedFormer consistently exceeds the performance of SAC at each epoch, robust to standard error. Focusing on the final epoch, we also obtain 1.56x performance gains in the \texttt{drawer-open} task, 1.36x performance gains in the \texttt{button-press} task, and 1.58x performance gains in the \texttt{window-open} task. In each other task, we obtain similar performance to SAC. Recall that SAC agents are centralized and train on the entire set of environments without preserving privacy, whereas FedFormer abides by the privacy constraints of federated learning with a distributed pool of agents. We attribute many of these gains to our transformer aggregation network, which can contextually leverage the individual expertise of agents without forcing a centralized model. 

\subsection{Scalability}
\label{sec:scalability}
\begin{figure*}
    \centering
    \includegraphics[width=\linewidth]{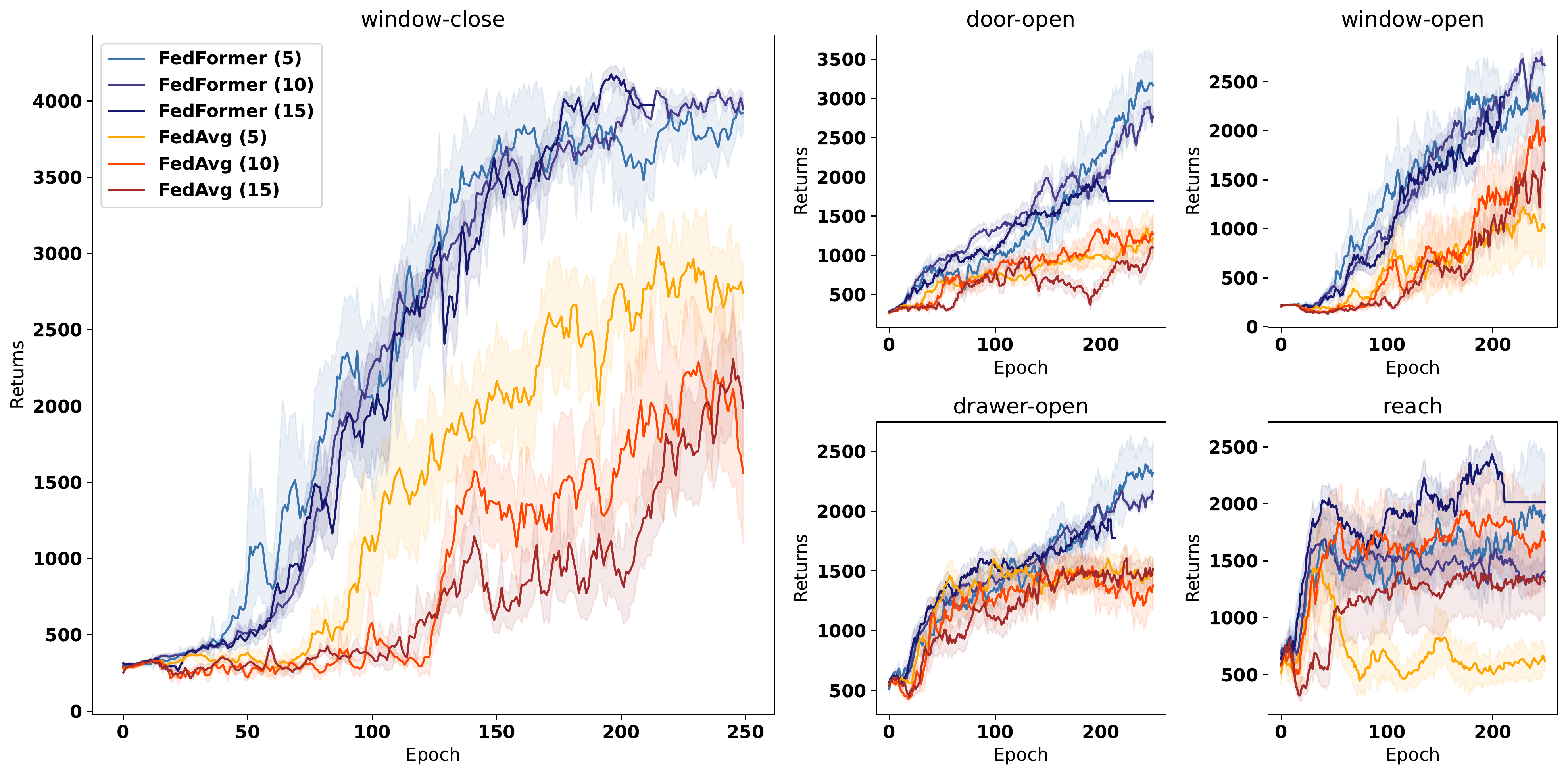}
    \caption{Performance when scaling federation methods to 5, 10 and 15 agents}
    \label{fig:scaling}
\end{figure*}

Next, we measure the ability of our method to scale to larger sets of agents. To do this, we compare the results of both federation strategies (FedAvg and our FedFormer) when scaled to 5, 10 and 15 agents (Figure \ref{fig:scaling}) \footnote{\revise{Although we evaluate our methods up to 15 agents, we can scale up to 512 agents (or 1024 with sparse variants) due to our usage of transformers}}. As was the case in our tests in Section \ref{sec:performance}, each agent is given five sampled environments for training and five sampled environments for testing without replacement. Since Meta-World generates 50 environments per task, some agents will have overlapping tasks for training that will be testing tasks for others, but never in the same agent. It is assumed that including more agents in federated learning should result in more robust models due to the larger availability of training data. However, it is also important to note that the performance reported is the average performance of \textit{all} agents. Therefore, all agents must perform well in order to achieve a high metric in our results. It is important to note that scaling poses a more difficult problem due to the increased heterogeneity of the environments that each agent is tasked with.

We find that increasing the number of agents results in better performance in FedFormer and worse performance for FedAvg. Focusing on the final epoch and when utilizing 10 agents, we can see performance improvements of 2.23x (\texttt{window-close}), 2.95x (\texttt{door-open}), 1.47x (\texttt{drawer-open}), 1.83x (\texttt{reach}) and 2.70x (\texttt{window-open}) compared to FedAvg. We see similar gains with 15 agents, where we can see performance improvements of 1.98x (\texttt{window-close}), 2.75x (\texttt{door-open}), 1.62x (\texttt{drawer-open}), and 1.88x (\texttt{window-open}). We also see that FedAvg fails to meaningfully scale above FedAvg (5), with FedAvg (5) and FedAvg (10) performing comparably or worse across all tasks. Many of the results reported are also robust to standard error. We hypothesize the poor performance of FedAvg is due to many agents attempting to explore the environment simultaneously but subsequently being forced into a single representation. \revise{In our experimentation, we found FedWeightedAvg to exhibit similar trends as FedAvg with marginal improvements as the number of agents increased.}

As a result, we conclude that our federation strategy can efficiently scale to include multiple agents. This result aligns with previous work in transformer models, which has shown the ability to reason over larger sets of embeddings \citep{devlin2018bert}.

\subsection{Agent Onboarding}
\label{sec:onboarding}
\begin{figure}
    \centering
    \includegraphics[width=\linewidth]{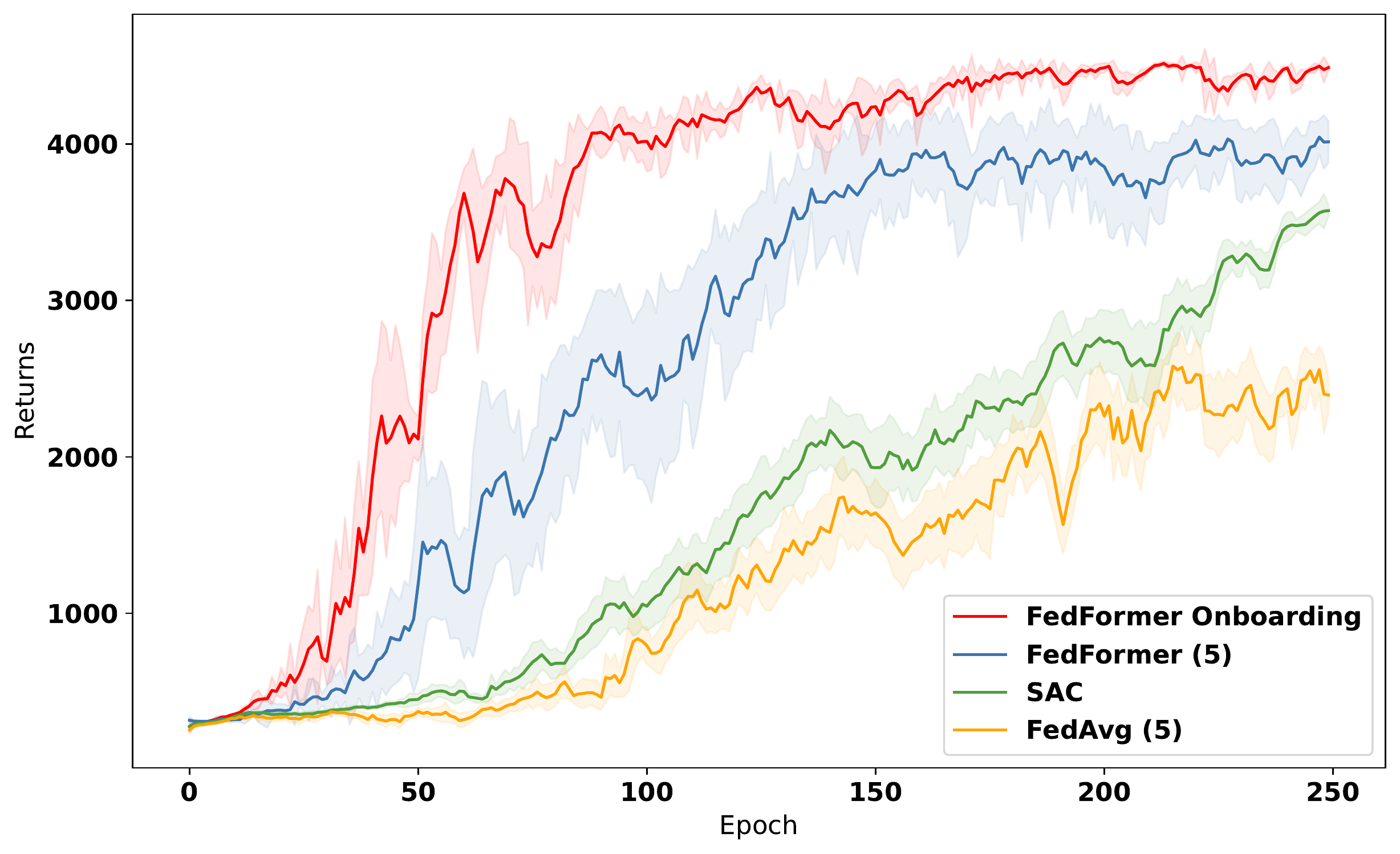}
    \caption{Performance of onboarding agents into pre-trained federation networks \revise{on the \texttt{window-close} task}}
    \label{fig:onboarding}
\end{figure}

In real-life use cases, agents are often introduced at random intervals into federation strategies. When an agent joins an established federation network, the network must adapt to account for the new environments introduced by the new agent. In the case of FedAvg, this can be a significant setback as model averaging can introduce noise into existing converged models \revise{when the models originate from heterogeneous environments}. In addition, FedAvg can require all agents to restart training to keep up with the modifications of the new agent. 

We now evaluate the onboarding strategy of FedFormer, which overcomes all these challenges (Section \ref{sec:arch}). To do this, we train a FedFormer network on the \texttt{window-close} task and save the resulting converged encoder networks. We then train a single new agent \edit{on five never-before-seen environments} utilizing the saved encoder networks of the previous agents as the external network set. The main difference here is that we do not update the external networks throughout training by the federation. In our experiments, we evaluate our methods by utilizing the resulting trained networks from each seed of the \texttt{window-close} task. The results for this experiment compared against FedFormer, SAC and FedAvg can be seen in Figure \ref{fig:onboarding}.

We see that the onboarded agent (\edit{red}) drastically benefits from the pre-trained encoder networks. Within the first 60 epochs, the new agent obtains near the equivalent performance of the original FedFormer network after it had trained for 250 epochs, obtaining a $\approx$ 5x efficiency increase. We also obtain 6.5x performance gains over SAC and 11.23x performance gains over FedAvg at epoch 60, demonstrating the increased speed of convergence. In addition, performance continues to improve beyond this point, obtaining a 1.24x increase over FedFormer at epoch 250. It is important to note that this onboarding process was done without any further training from the other agents.  

\subsection{Ablation Study}
\begin{figure*}
    \centering
    \includegraphics[width=\linewidth]{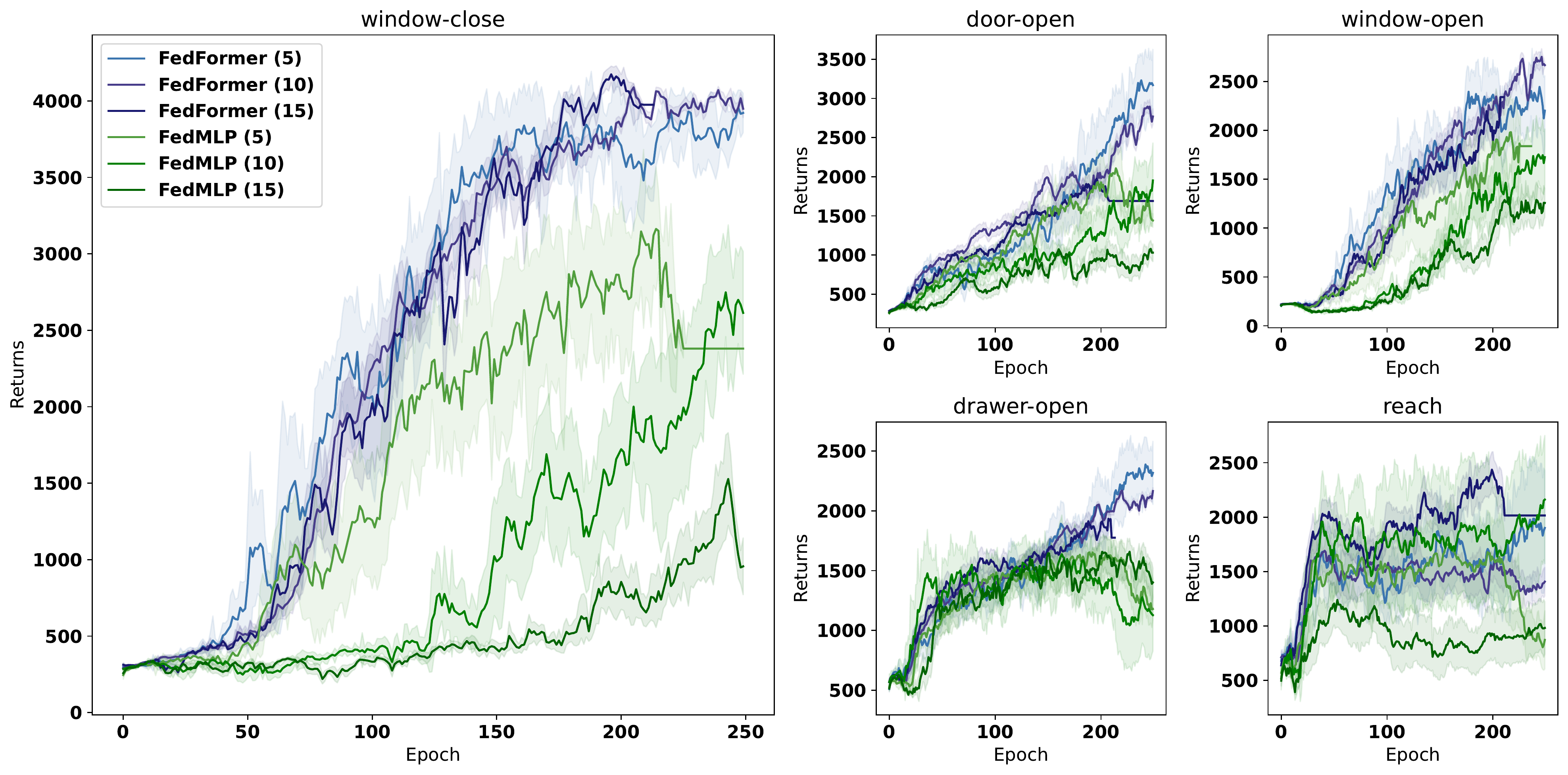}
    \caption{Scaling Performance when replacing Transformer Network with MLP Network}
    \label{fig:ablation}
\end{figure*}
\edit{To evaluate the usefulness of attention in our federation network, we have conducted additional experiments replacing the Transformer network in FedFormer with a three-layer non-attention multi-layer perceptron model (FedMLP). To do this, we concatenate the embeddings generated by each of the encoder networks as the input to the MLP. Then, the output layer of the MLP generates an embedding of size 256, matching the dimensionality of the CLS embeddings created by FedFormer. Just like FedFormer, this embedding is then decoded by a local decoder network to generate the respective Q value for that state-action pair (Figure \ref{fig:arch}).} \revise{This baseline is similar to work by \citet{zhuo2019federated}, which proposed concatenating the Q-network embeddings of two agents into a global MLP.} Our results when scaling this method from 5, 10 and 15 agents can be seen in Figure \ref{fig:ablation}, following the procedure outlined in Section \ref{sec:scalability}. It is important to note that scaling poses a more difficult problem due to the increased heterogeneity of the environments that each agent is tasked with. 

In each of the tasks evaluated, we see the dominant performance of FedFormer over the FedMLP baseline. This can be most evidently seen in the \texttt{window-close} task, where FedFormer (5) often achieves over 2x the performance of FedMLP (5) across each epoch. It is important to note the large differences between the two methods when scaled to include larger agent pools. The largest difference can again be seen on the \texttt{window-close} task, where the performance of FedMLP drops by 2.4x when scaled to 10 agents and 3.8x when scaled to 15 agents. This is contrasted by FedFormer, where the performance is either maintained or increased with increased agents. These results indicate the importance of the attention mechanism for FedFormer, especially in relation to scalability. With attention, it is possible to contextually select and aggregate the insights of participating agents in relation to the needs of the current agent. This is contrasted by FedMLP, which must aggregate the insights of all agents, including agents that provide irrelevant input. The problem of noisy inputs is then further exacerbated with increased agent counts, as seen in Figure \ref{fig:ablation}.

\section{Related Work}
\label{sec:related}

A similar field to federated reinforcement learning is multi-agent reinforcement learning (MARL). MARL focuses on tasks where multiple agents interact in the same environment and share insights with each other to solve a cooperative or competitive task \citep{stone2000multiagent}. An important challenge with these systems is the non-stationarity of the environment, where each agent has their own perspective on the environment and actions committed by a single agent can affect the environment for another \citep{zhang2021multi}. Therefore, it becomes critical that multiple agents can efficiently share information, similar to federated learning \citep{foerster2016learning, sukhbaatar2016learning, mao2018Modelling}. However, MARL allows for the constant flow of observation information between agents and often uses a centralized model to plan the actions of each agent \citep{mao2018Modelling, han2019IPOMDP}, therefore making the methods unsuitable for federated learning.

One MARL approach related to our work is Multi-Agent-Attention-Critic (MAAC), proposed by \citet{iqbal2019actor}. This method proposed utilizing transformer models to coordinate cooperative agents to solve a collective task. During each timestep, each agent broadcasts their local observation-action pairs to a centralized model. These inputs are then aggregated using a transformer encoder into a set of Q-Values corresponding to each agent, which is then broadcast back to each agent. As a result, each resulting Q-value is created with respect to each other agent’s observations and the model is fine-tuned end-to-end to optimize a joint goal. 

While our approach also utilizes transformer models, there are significant differences, notwithstanding the obvious constraints of privacy. The transformer model in MAAC is used for cooperative learning, where the central model coordinates observation embeddings from many agents towards a shared goal. In our work, each agent has their own decentralized transformer network that is optimized for aggregating embeddings toward that agent's goal. In addition, the input to each agent is a set of embeddings encoding the same local observation but with different federated encoder networks, whereas MAAC requires observation sharing from each agent during each timestep. We also include elements from natural language processing, such as [CLS] tokens to encode global representations and position encodings \citep{devlin2018bert}.

\section{Conclusion}
\label{sec:future_work}
We presented a novel federation strategy based on transformer models, FedFormer. Our technique addresses concerns about utilizing federated learning towards reinforcement learning, which is typically done through the FedAvg method of averaging parameters. Instead, our method utilizes attention to compute contextual relationships between agents without compromising the ability of individual agents to explore the environment. 

Our method outperforms FedAvg and single-agent Soft Actor-Critic (SAC), with FedFormer outperforming FedAvg by a factor of 4.4 and SAC by a factor of 1.75. In addition, we demonstrate that FedFormer can effectively scale to more agents, with gains of 2.4x and 3.8x over FedAvg with 10 and 15 agents respectively. This contrasts with FedAvg, which has negligible performance gains and at times performs worse. Lastly, we also demonstrate the ease of agent onboarding in our methods, resulting in 5x convergence speeds by leveraging pre-trained models without sacrificing the performance of other models. All of this is achieved while still abiding by the privacy constraints of federated learning. 

\edit{The primary limitation of our work is related to computational latency. Previous work in transformer models states a computational limit of $\approx 512$ embeddings when computing attention \cite{devlin2018bert}. This is due to the quadratic increase in latency with additional embeddings. Further, each new agent will add an encoder network to the model. As a result, this limits the number of agents that can participate as part of the federation network. However, this limit can be addressed by utilizing domain-specific agent sampling strategies, such as those explored by \citet{jianji2019federated}, \citet{wang2020federated} and \citet{tehrani2021Federated}.}

With our advancements with introducing FedFormer, there are several future work directions that could be pursued. First, since our approach does not rely on averaging parameters, it would be interesting to explore the use of deep learning models to represent the encoder networks \citep{mnih2013playing}. This could potentially allow for the federation to tackle more complicated tasks involving computer vision and partial observability by leveraging deep encoders such as convolutional neural networks and recurrent neural networks. The utilization of these models would have been previously difficult due to instability brought on by parameter averaging \citep{qi2021federated}. 

\revise{Another interesting direction for future work is to explore related work in supervised heterogeneous federated learning. This area proposes techniques to group and filter agents, including to address tolerance to faults from byzantine agents \cite{gao2022heterogenous}. These techniques support learning from a set of changing heterogeneous environments but often assume that the environments (datasets) of individual agents remain consistent \cite{gao2022heterogenous}. While this is often not the case in federated reinforcement learning, these techniques still present interesting insights to explore. For example, \citet{Huang_2022_CVPR} could be adapted to cluster agents based on their Q-values, \citet{Fang_2022_CVPR} could be repurposed to address the detection of byzantine agents. The work of \citet{fan2021fault} and \citet{chen2022byzantine} will also become relevant. Byzantine filtering layers proposed by \citet{fan2021fault} could be applied on the input encoder networks; Chen's insights into theoretical guarantees may inspire us to expand beyond our current experimental validation. Since both \citet{fan2021fault} and \citet{chen2022byzantine} are grounded in FedAvg, using FedFormer instead with these approaches may provide important steps forward.}

Lastly, our work opens the door for advances in existing FedAvg-based architectures. From IoT devices to self-driving cars, we believe that our methods can serve as a direct stand-in for FedAvg to provide better performance. Further evaluation of these tasks with the addition of domain-specific adjustments such as those proposed in previous work \citep{jianji2019federated, wang2020federated, tehrani2021Federated, nadiger2019federated} would be especially interesting. We hope that our work can be used to improve federated reinforcement learning and introduce advancements in deep learning to this critical field, especially as privacy becomes harder to obtain in favour of performance.


\begin{acks}
\revise{The authors thank the Natural Sciences and Engineering Research Council of Canada, the Canada Research Chairs Program, the Vector Institute and the University of Waterloo Cheriton Scholarship for financial support. We are also grateful to the reviewers for their valued feedback on the paper.}
\end{acks}



\bibliographystyle{ACM-Reference-Format} 
\bibliography{sample}


\appendix

\end{document}